\documentclass[letterpaper, 10 pt,conference]{ieeeconf}
\IEEEoverridecommandlockouts
\let\labelindent\relax

\usepackage{caption}
\usepackage{tikz, tikz-3dplot}

\graphicspath{{figures/}} 
\DeclareGraphicsExtensions{.pdf,.jpg,.png} 
\usepackage{amsmath} 
\usepackage{amsfonts} 
\usepackage{tikz} 
\usetikzlibrary{shapes,arrows,fit,calc,positioning,automata}
\usepackage{dblfloatfix} 
\usepackage{color} 
\usepackage{multirow} 
\usepackage{hhline} 
\usepackage[bookmarks=false]{hyperref} 
\usepackage{enumerate} 
\usepackage{epstopdf} 
\usepackage{todonotes}
\usepackage{svg}
\usepackage{pdfpages}
\usepackage{algorithm}
\usepackage[export]{adjustbox}
\usepackage{graphicx}
\usepackage{setspace}
\usepackage{graphicx}
\usepackage{caption}
\usepackage{subcaption}
\usepackage{amsmath}
\usepackage{amssymb}
\usepackage{algorithm}
\usepackage{algorithmic}
\usepackage{xcolor}
\usepackage{booktabs}
\usepackage{multirow}
\usepackage[table]{colortbl}
\usepackage{diagbox}
\usepackage{makecell}
\usepackage{enumitem}
\usepackage{tikz}
\usepackage[most]{tcolorbox}
\usepackage[nolist]{acronym}
\usepackage{balance}

\begin{acronym}[MPC]
\acro{GSD}{ground sample distance}
\acro{UAV}{unmanned aerial vehicle}
\acrodefplural{UAV}{unmanned aerial vehicles}
\acro{SGD}{stochastic gradient descent}
\acro{GPS}{global positioning system}
\acrodefplural{GPS}{global positioning system's}
\acro{CNN}{convolutional neural network}
\acro{HarDNet}{harmonic densely connected network}
\acro{DDRNet}{deep dual-resolution networks}
\acro{BiSeNet}{bilateral segmentation network}
\acro{DNN}{deep neural network}
\acrodefplural{DNN}{deep neural networks}
\acro{FOV}{field of view}
\acro{DCNN}{deep convolutional neural network}
\end{acronym}

\begin{document}

%
\title{\LARGE \bf Safe Vessel Navigation Visually Aided by Autonomous Unmanned Aerial Vehicles in Congested Harbors and Waterways }

\author{ Jonas le Fevre Sejersen, Rui Pimentel de Figueiredo and Erdal Kayacan
\thanks{J. Fevre, R. Figueiredo, E. Kayacan are with Artificial Intelligence in Robotics Laboratory (Air Lab), the Department of Electrical and Computer Engineering, Aarhus University, 8000 Aarhus C, Denmark
        {\tt\small \{jonas.le.fevre,rui,erdal\} at ece.au.dk}}%
}


%


\maketitle







\begin{abstract}
    In the maritime sector, safe vessel navigation is of great importance, particularly in congested harbors and waterways. The focus of this work is to estimate the distance between an object of interest and potential obstacles using a companion UAV. The proposed approach fuses \ac{GPS} data with long-range aerial images. First, we employ semantic segmentation \acp{DNN} for discriminating the vessel of interest, water, and potential solid objects using raw image data. The network is trained with both real and images generated and automatically labeled from a realistic AirSim simulation environment. Then, the distances between the extracted vessel and non-water obstacle blobs are computed using a novel \ac{GSD} estimation algorithm.  To the best of our knowledge, this work is the first attempt to detect and estimate distances to unknown objects from long-range visual data captured with conventional RGB cameras and auxiliary absolute positioning systems (e.g. \ac{GPS}). The simulation results illustrate the accuracy and efficacy of the proposed method for visually aided navigation of vessels assisted by \acp{UAV}.
\end{abstract}

\IEEEpeerreviewmaketitle


\section{Introduction}
\label{sec:intro}


In the maritime sector, request for accurate and safe vessel navigation induces a need for reliable operation due to high risks of accidents when maneuvering in highly dynamic environments. In this study, an unmanned aerial vehicle (UAV)-assisted vessel navigation framework is proposed. The focus of this work is  to estimate the distance between an object of interest (i.e. a vessel) and potential obstacles (i.e., shipyard objects), using a companion UAV.


There are several successful robotics and autonomous driving methods in which relative locations and distances are accurately determined for successful and safe interaction with the surrounding environment. However, to the best of our knowledge, the literature lacks methods to determine relative distances between an object of interest and unknown objects in the environment.
\begin{figure}[t]
  \centering
  \includegraphics[keepaspectratio, width=0.48\textwidth]{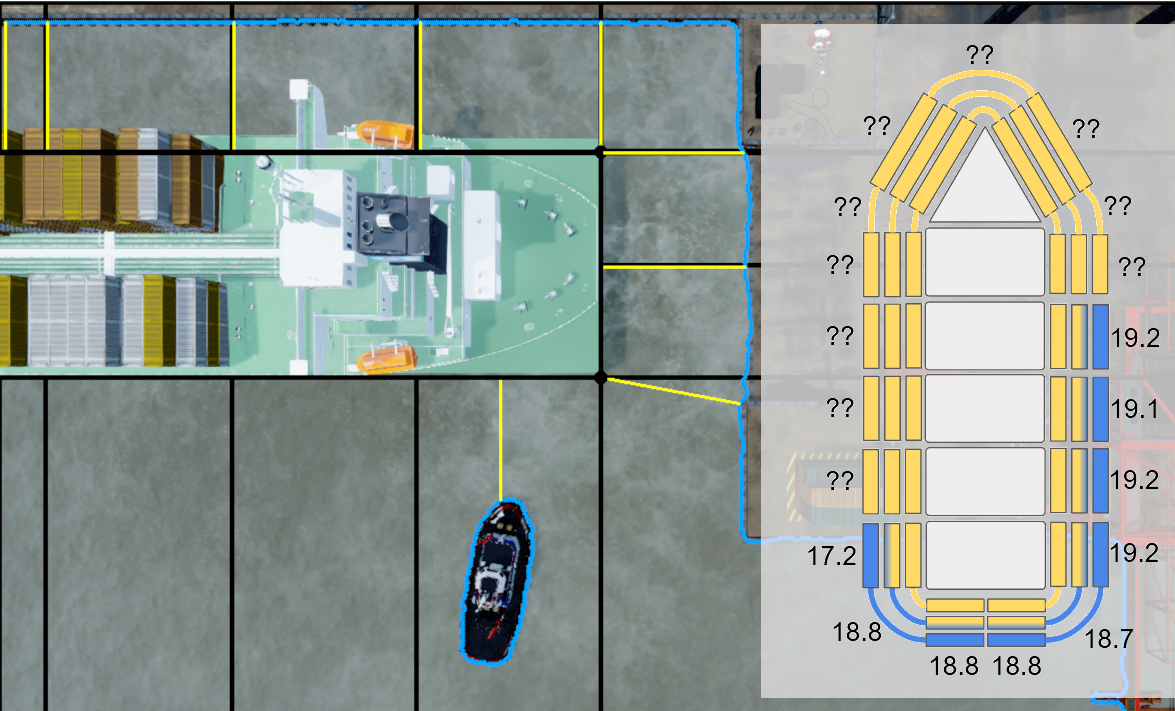}
  \caption{On the left we see each section of the vessel, and their correspondent region (squares in black). The estimated shortest distance is noted as a yellow line going from the object and projected into the corresponding section on the vessel. A distance measure is associated to each section of the vessel, and its closest obstacle. Sections with no close-by objects within the field-of-view are annotated with two question marks.}
  \label{fig:method_output} 
\end{figure}
In this study, we propose a long-range aerial obstacle detection methodology for vessel guidance. Our solution combines absolute positioning measurements given by \ac{GPS} and magnetometer devices, located on the vessel and a companion \ac{UAV}, with visual information provided by a consumer-grade RGB camera mounted on the bottom of the \ac{UAV}.
The contributions of this study are the followings:
\begin{itemize}
    \item First, we employ the state-of-the-art \ac{CNN}-based semantic segmentation methods for real-time extraction of complex appearance and highly-variable vessel, water, and potential non-water obstacle silhouettes in pixel coordinates. We create a large annotated dataset using a realistic AirSim environment for training and testing the semantic segmentation performance on shipyard structures. 
    \item Second, we propose a method using absolute positioning data provided by \ac{GPS} and magnetometers located on the vessel and \ac{UAV} to accurately determine where the object of interest (i.e. a vessel) is located from 3D world coordinates to the image plane. A \ac{GSD} strategy is employed to determine relative distances between objects on the ground.
    \item Finally, we thoroughly assess our methods with a set of experiments in a realistic simulation scenario and with real data. The experiments demonstrate that our real-time approach is accurate and suitable for range estimation from aerial images and absolute positioning data provided by \ac{GPS} and magnetometer.
\end{itemize}


The rest of this article is organised as follows: in Section \ref{sec:RelatedWork} we overview the state-of-the-art on image-based semantic segmentation, localization and range estimation. In Section \ref{sec:metho} we describe the proposed solution for computing distance from aerial images acquired from long distance systems (e.g. flying drone). Lastly we wrap-up with results in Section \ref{sec:ex_res}, while drawing ideas for future work in Section \ref{sec:discussion}.

\section{Related Work}
\label{sec:RelatedWork}
\subsection{Image Semantic Segmentation}
Semantic segmentation deals with the problem of associating image pixel values to object classes and has become an important topic in the last few years in various domains. Particularly in biomedical and robotics applications, with impressive achievements in autonomous car driving applications. 
In this section, we overview different neural network-based methods, which are the top performers in visual semantic segmentation tasks, including the large-scale Cityscapes dataset and bench-marking suite for visual semantic scene understanding \cite{Cordts_2016_CVPR}. 

The \ac{BiSeNet} \cite{bisenet2018} comprises two different paths: the spatial path, which encodes low-level high detailed information, and the context path, which mainly encodes high-level context information. A feature fusion module is used to fuse features from these two paths. First, these are concatenated, and then batch normalization is used to balance the scale of the features. Finally, the concatenated features are pooled and re-weighted using a weight vector. 

The \ac{HarDNet} \cite{Chao_2019_ICCV} is a sparse, highly efficient, low memory traffic \ac{CNN}, based on the sparse densely connected networks (DenseNets) \cite{dense_networks2017}.  Unlike traditional \acp{CNN} with $L$-layers that have $L$ connections, in \ac{HarDNet} a layer $k$ creates a connection to layer $k - 2^n$ if $2^n$ divides $k$, where $n$ is a non-negative integer and $k-2^n \geq 0$. Utilizing this connection method, once layer $2^n$ is processed, each layer from 1 to $2^n - 1$ can be flushed from memory. They significantly reduce the number of parameters, alleviate the vanishing-gradient problem, and improve feature reuse along layers. Hong et al. \cite{Chao_2019_ICCV}  propose removing layers with very low Model of Computation (MoC), reducing DRAM traffic by $40\%$ when compared with DenseNets \cite{dense_networks2017}.

The authors of \cite{hong2021deep} propose a \ac{DDRNet} for real-time semantic segmentation of road scenes. The main novelty is a contextual information extractor named Deep Aggregation Pyramid Pooling Module (DAPPM) that simultaneously enlarges receptive fields and fuses multi-scale context. When compared to the former, \ac{DDRNet} promotes more information sharing across layers and the generation of higher-resolution feature maps with sufficiently large receptive fields.

The use of neural network-based approaches has been successfully employed in various drone applications including surveillance \cite{ilkeriros2020,bozcancontext} and path planning \cite{ecamci2020, camci2019endtoend}. In this work, we utilize this to tackle the problem of discriminating objects from water surfaces.

\subsection{Range Measurement Techniques}
Different techniques exist in the literature for extracting the location between objects in the environment, using a variety of sensors. On one hand, methods based on sensors that capture the internal dynamics of the system, such as accelerometers, magnetometers, \acp{GPS}, encoders, are often referred to as being proprioceptive-based methods. On the other hand, methods based on contactless sensors, in particular Lidar, RGB-D, Radar, or a combination of the former, are often called exteroceptive-based methods. Proprioceptive sensors are typically suitable for self-localization and motion estimation, whereas the latter are more indicated for the perception of the surrounding environment, e.g. determining relative distances to external objects in the environment.

In the work of \cite{lidarCameraFusion2020}, the authors present an approach for estimating the distance between self-driving cars and other vehicles and objects along its path, using high precision exteroceptive lidar and RGB data \cite{measuringDistanceRadar2019}. In \cite{8356033}, the authors use a 2D laser scanner to assess the evenness of walls in a service robot application. In \cite{deepLearningVehicle2020} the authors provide a thorough overview of sensor fusion methods, both for proprioceptive and exteroceptive data that were successfully employed in autonomous driving applications.
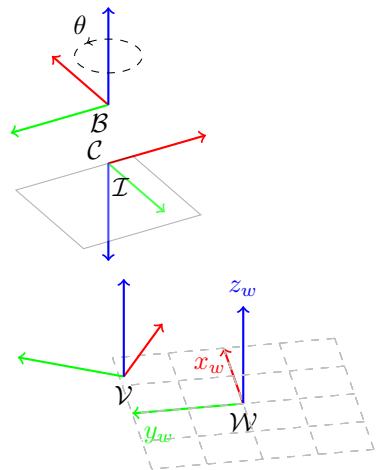
\begin{figure}[b!]

\tdplotsetmaincoords{60}{240}
\pgfmathsetmacro{\rvec}{.8}
\pgfmathsetmacro{\thetavec}{30}
\pgfmathsetmacro{\phivec}{60}
\pgfmathsetmacro{\axisscale}{0.5}

\pgfkeys{%
/polargrid/.cd,
rmin/.code ={\global\def\rmin {#1}},
rmax/.code ={\global\def\rmax {#1}},
amin/.code ={\global\def\amin {#1}},
amax/.code ={\global\def\amax {#1}},
rstep/.code={\global\def\rstep{#1}}, 
astep/.code={\global\def\astep{#1}}}

\def\polargrid{\pgfutil@ifnextchar[{\polar@grid}{\polar@grid[]}}%
\def\polar@grid[#1]{%
\pgfkeys{/polargrid/.cd,
rmin ={0},
rmax ={0.5},
amin ={-90},
amax ={90},
rstep={0.09}, 
astep={22.5}}   
\pgfqkeys{/polargrid}{#1}%
\pgfmathsetmacro{\addastep}{\amin+\astep} 
\pgfmathsetmacro{\addrstep}{\rmin+\rstep} 
 \foreach \a in {\amin,\addastep,...,\amax}  \draw[tdplot_rotated_coords,gray] (\a:\rmin) -- (\a:\rmax);  
 \foreach \r in {\rmin,\addrstep,...,\rmax}  \draw[tdplot_rotated_coords,gray] (\amin:\r) arc (\amin:\amax:\r);    
 }

\pgfmathsetmacro{\arcradius}{0.15}

\makeatletter
\tikzoption{canvas is xy plane at z}[]{%
  \def\tikz@plane@origin{\pgfpointxyz{0}{0}{#1}}%
  \def\tikz@plane@x{\pgfpointxyz{1}{0}{#1}}%
  \def\tikz@plane@y{\pgfpointxyz{0}{1}{#1}}%
  \tikz@canvas@is@plane
}
\makeatother

\tikzset{xyp/.style={canvas is xy plane at z=#1}}
\tikzset{xzp/.style={canvas is xz plane at y=#1}}
\tikzset{yzp/.style={canvas is yz plane at x=#1}}



    \centering

\tikzstyle{background grid}=[draw, black!50,step=.1cm]    
\begin{tikzpicture}[scale=3,tdplot_main_coords]

\coordinate (O) at (0,0,0);

\pgfmathsetmacro{\focaldistance}{0.2}
\pgfmathsetmacro{\opticalaxislength}{2.1}
\pgfmathsetmacro{\planexlength}{0.3}
\pgfmathsetmacro{\planeylength}{0.3}
\pgfmathsetmacro{\pan}{0}
\pgfmathsetmacro{\tilt}{0}
\pgfmathsetmacro{\vergenceangle}{22}

\tdplotsetcoord{P}{\rvec}{\thetavec}{\phivec}

\tdplotsetcoord{WORLD}{0.6}{90}{-125}
\tdplotsetcoord{GPS}{0.2}{90}{-50}
\tdplotsetcoord{PTU}{1.5}{0}{90}
\tdplotsetcoord{EGO}{1.55}{0}{90}

\tdplotsetcoord{LCO}{2.0}{40}{38+\pan}
\tdplotsetcoord{CAM}{1.2}{0}{90}
\tdplotsetcoord{CYC}{1.8}{30}{+\pan}


\tdplotsetrotatedcoords{-20}{0}{0}
\tdplotsetrotatedcoordsorigin{(WORLD)}
\draw[tdplot_rotated_coords]  (0,0,0) node[anchor=north] {$\mathcal{W}$};
\draw[thick,tdplot_rotated_coords,->,red] (0,0,0) -- (\axisscale,0,0) node[anchor=north west,xshift=-1.5em]{$x_{w}$};
\draw[thick,tdplot_rotated_coords,->,green] (0,0,0) -- (0,\axisscale,0) node[anchor=east,xshift=1.95em,yshift=-0.8em]{$y_{w}$};
\draw[thick,tdplot_rotated_coords,->,blue] (0,0,0) -- (0,0,\axisscale) node[anchor=south]{$z_{w}$};

\foreach \x in {-0.5,-0.25,...,0.5}
    \foreach \y in {-0.5,-0.25,...,0.5}
    {
        \draw[very thin,gray!50,dashed,tdplot_rotated_coords] (\x,-0.5,0) -- (\x,0.5,0);
        \draw[very thin,gray!50,dashed,tdplot_rotated_coords] (-0.5,\y,0) -- (0.5,\y,0);
    };
    
    
\tdplotsetrotatedcoords{-50}{0}{0}
\tdplotsetrotatedcoordsorigin{(GPS)}
\draw[tdplot_rotated_coords]  (0,0,0) node[anchor=north] {$\mathcal{V}$};
\draw[thick,tdplot_rotated_coords,->,red] (0,0,0) -- (\axisscale,0,0);
\draw[thick,tdplot_rotated_coords,->,green] (0,0,0) -- (0,\axisscale,0);
\draw[thick,tdplot_rotated_coords,->,blue] (0,0,0) -- (0,0,\axisscale);

\tdplotsetrotatedcoords{\pan}{\tilt}{0}
\tdplotsetrotatedcoordsorigin{(PTU)}
\draw[tdplot_rotated_coords]  (0,0,0) node[anchor=north,xshift=-0.35em] {$\mathcal{B}$};
\draw[thick,tdplot_rotated_coords,->,red] (0,0,0) -- (\axisscale,0,0);
\draw[thick,tdplot_rotated_coords,->,green] (0,0,0) -- (0,\axisscale,0);
\draw[thick,tdplot_rotated_coords,->,blue] (0,0,0) -- (0,0,\axisscale);

\draw[dashed,tdplot_rotated_coords,xyp=\axisscale/2,->] (\arcradius,0) arc (0:370:\arcradius) node [anchor=south,xshift=0.5em,xshift=-0.75em] {$\theta$};


\tdplotsetrotatedcoords{\pan}{180+\tilt}{-90}
\tdplotsetrotatedcoordsorigin{(CAM)}
\draw[tdplot_rotated_coords]  (0,0,0) node[anchor=north,yshift=1.2em,xshift=-0.5em] {$\mathcal{C}$};

\draw[thick,tdplot_rotated_coords,->,red] (0,0,0) -- (\axisscale,0,0);
\draw[thick,tdplot_rotated_coords,->,green] (0,0,0) -- (0,\axisscale,0);
\draw[thick,tdplot_rotated_coords,->,blue] (0,0,0) -- (0,0,\axisscale);



\begin{scope}[tdplot_rotated_coords,xyp=\focaldistance]
    \fill[opacity=0.3,white,draw=black] (-\planexlength,-\planeylength) rectangle (\planexlength,\planeylength);
    \fill[black] (0.1,-0.1) node[right, scale=1.0, xshift=-0.5em] {$\mathcal{I}$};
\end{scope}    

\end{tikzpicture}
\caption{Coordinate reference frames used by our system (best seen
in color): the inertial world frame $(\mathcal{W})$ in which the environment is represented; the camera frame $(\mathcal{C})$, in which monocular images are obtained, which is rigidly attached to the \ac{UAV} robot base $(\mathcal{B})$, and permits determining the \ac{UAV} pose in the world, given visual measurements and absolute positioning readings given from \acp{GPS} located in the robot base and vessel ($\mathcal{V}$).}
\label{fig:coordinate_frames}
\end{figure}

Vision-based approaches typically rely on a combination of RGB and depth information; however, depth sensors become less accurate as the distance grows, being unsuitable for long-range distance estimation when object or landmark locations are unknown. This is typically the case when one needs to detect or measure distances of objects on the ground \cite{ilker2021},\cite{audataset2020}, using aerial images captured from \acp{UAV} or satellites. \ac{GSD} techniques \cite{felipe2012analysis} are often employed to tackle this problem by determining the size of each on the ground in metric units.

\section{Methodologies}
In this work, we combine proprioceptive data given by \ac{GPS}, and magnetometer sensors located on the target vessel and companion drone, with exteroceptive visual data provided from a downward-facing camera located on the drone, to compute absolute distances between the vessel surface and obstacles in the environment.

\subsection{Problem Definition}
\label{sec:metho}
The problem tackled in this work is: given an aerial RGB image and the global pose of the \ac{UAV} and the object of interest (i.e., vessel), estimate the distance between the vessel and the surrounding obstacles, to be used as an auxiliary guidance mechanism for safe navigation in harbor scenarios.

\begin{figure*}[ht]
    \centering
    \includegraphics[width=0.99\textwidth]{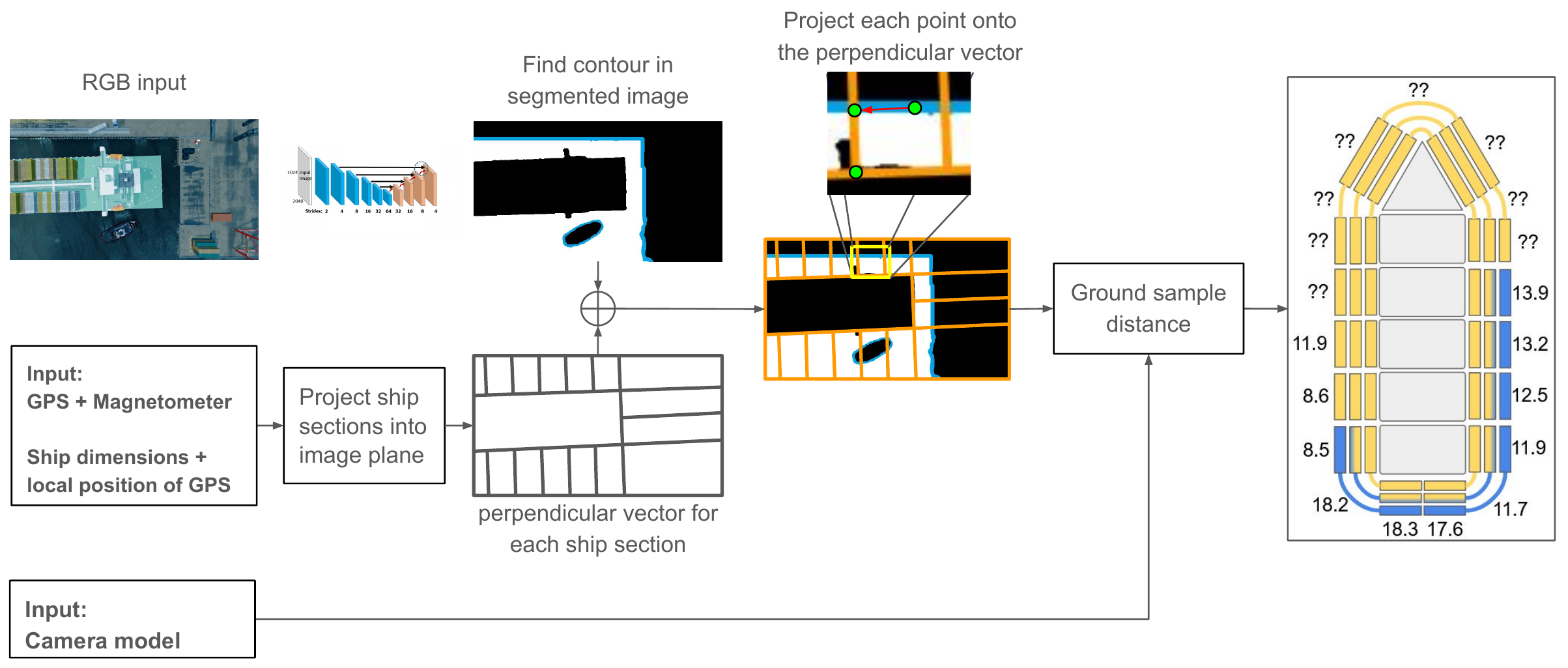}
    \caption{Proposed method for inter-object distance computation from aerial image and proprioception data.}
    \label{fig:proposed_pipeline}
\end{figure*}

The proposed method (Fig. \ref{fig:proposed_pipeline}) first relies on a semantic segmentation approach for extracting locations of objects and vessels in the environment that receives RGB images as input. The network first performs multi-class segmentation (i.e. water, ship, dock object). Then it binarizes the image to distinguish between solid objects (i.e. ships and dock objects) and water in order to extract contours that enclose solid structures in the shipyard environment. Finally, the proposed method computes the distance between lines belonging to the vessel of interest (distinguished from other vessels using an onboard \ac{GPS}) and every other object in the environment (i.e. other vessels and dock objects).

\subsection{System Overview}

The adopted environment structure representation, i.e. 3D map, is a discrete projection of the world structure $W\subset\mathbb{R}^3$ in the world reference frame $\mathcal{W}$. In order to be able to perform absolute distance measuring, the proposed system is equipped with two sensing modalities: 
\begin{itemize}
    \item proprioceptive - absolute world positioning ($\mathcal{W}$) from \ac{GPS}, magnetometer and barometer sensors located both in the vessel and \ac{UAV}
    \item exteroceptive - visual recognition abilities and relative distances between objects in the environment provided by a monocular camera installed on the \ac{UAV}.
\end{itemize}
At each time instant, the proprioceptive modalities report the \ac{UAV} and vessel state. More specifically, the global positions and headings of the \ac{UAV} and vessel, $X^\mathcal{B}, X^\mathcal{V}\in\mathbb{R}^6$, respectively, in the world inertial reference frame $\mathcal{W}$. The transformation from the vessel ($\mathcal{V}$) to camera pixel coordinates ($\mathcal{I}$) can be recovered applying the following camera projective model
\begin{equation}
    T^\mathcal{I}_\mathcal{V}=K T^\mathcal{C}_\mathcal{B}T^\mathcal{B}_\mathcal{W}T^\mathcal{W}_\mathcal{V}
\end{equation}
where $K$ represents the camera intrinsic parameter matrix and $T$ the transformations between frames. For the sake of simplicity, we assume the transformation between the camera ($\mathcal{C}$) and the \ac{UAV} base ($\mathcal{B}$) is fixed, and the other transformations determined from the noise-free proprioceptive modalities. In other words, we consider that the positioning measurement errors are negligible, and therefore the transformations between the various reference frames involved in our system (see Fig. \ref{fig:coordinate_frames}) can be deterministic determined. 

At each time instant, the sensing system computes the distance between the vessel and the closest obstacle found, given the list of distance estimates defined in the camera optical reference frame $\mathcal{C}$, estimated from RGB images, and noisy proprioception data. The observation model described in Section \ref{sec:observation_model} explains how distance measurements $Z_t$ are generated from the environment given \ac{GPS} and visual RGB data only.

\subsection{Observation Model}
\label{sec:observation_model}
The proposed method relies on \ac{GPS} and magnetometer to simultaneously estimate the position and attitude of the \ac{UAV} and the vessel of interest. To distinguish between solid obstacles and safe navigation areas (i.e. water), we begin by first employing a semantic segmentation algorithm that masks out image regions belonging to water. The remaining regions not belonging to the vessel of interest are considered potential objects used for computing distances.
\begin{figure}[t]
  \centering
  \begin{subfigure}[b]{0.48\textwidth}
      \includegraphics[width=1.0\textwidth]{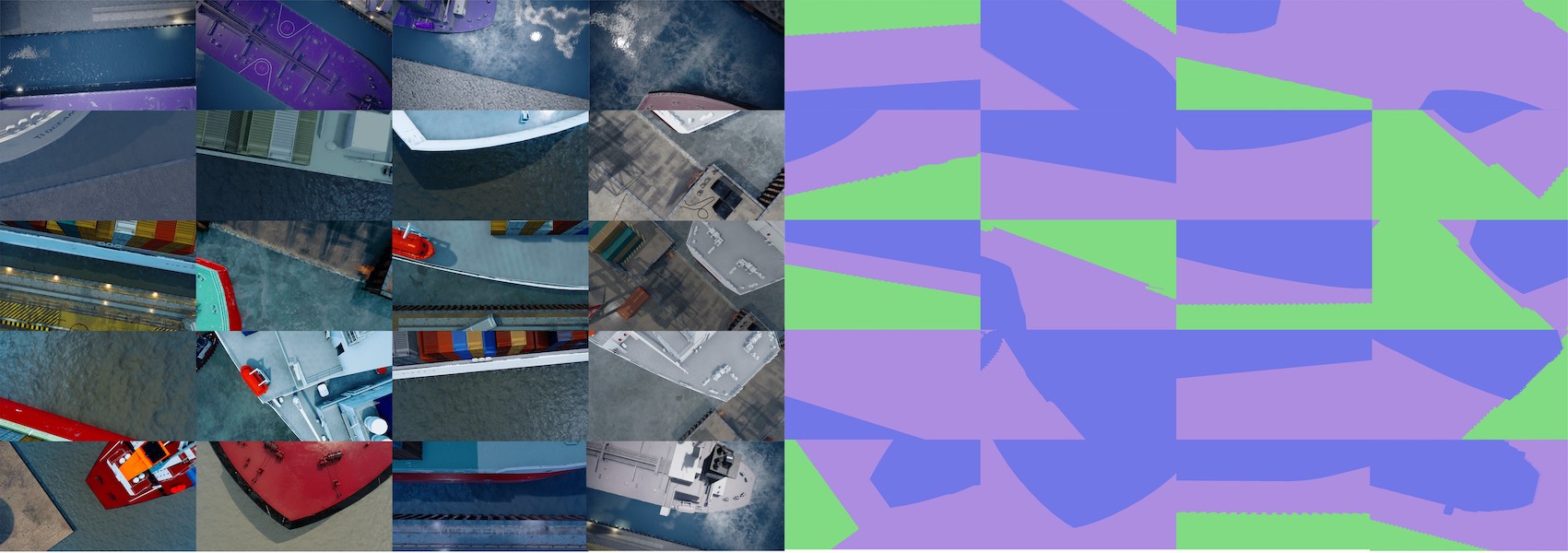}
      \caption{}
  \end{subfigure}
  \begin{subfigure}[b]{0.48\textwidth}
      \includegraphics[width=1.0\textwidth]{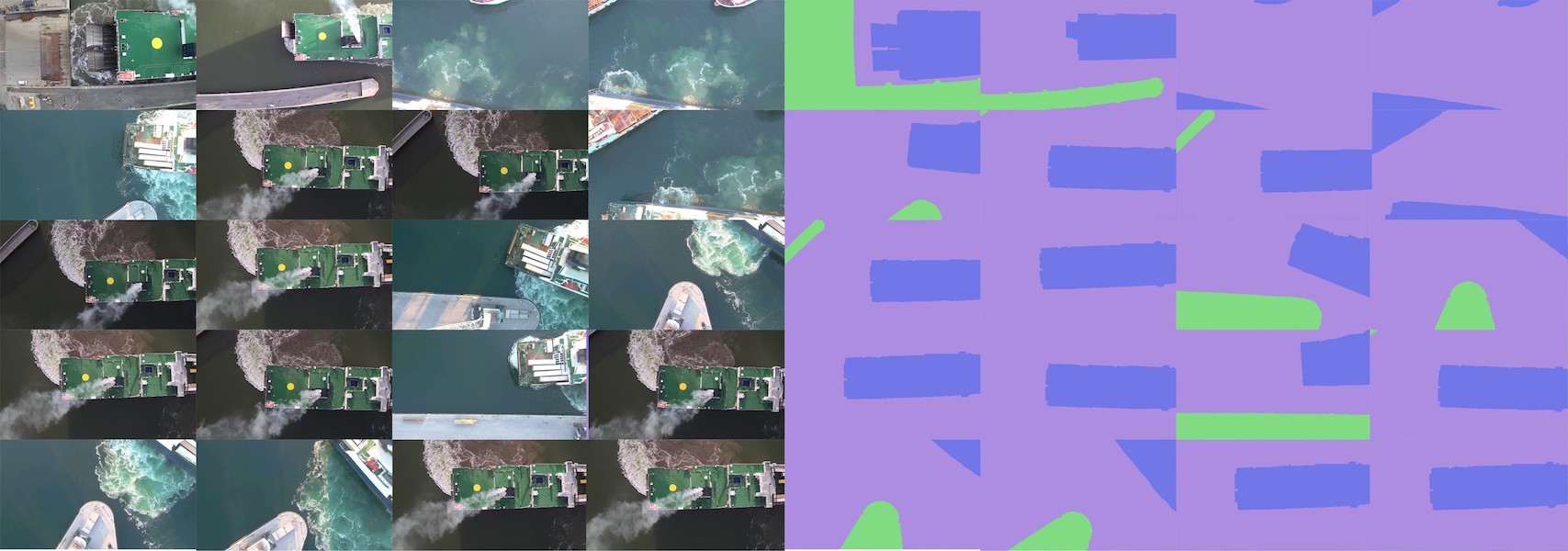}
      \caption{}
  \end{subfigure}

  \caption{Example AirSim (a) and real (b) labeled datasets.}
  \label{fig:example_labeled_images} 
\end{figure}

\subsubsection{Image-based Semantic Segmentation}
In order to detect shipyard obstacles and vessel structures, we rely on image data provided by the pointing bottom camera located on the bottom of the \ac{UAV}, and data-driven semantic segmentation methods based on \acp{DNN}, that provide pixel-level class probability distributions regarding the likelihood of each class, pertaining in RGB images. These \acp{DNN} are optimized with large annotated image data-sets, acquired using realistic physics and rendering engines. We have tested three different networks \cite{bisenet2018,Chao_2019_ICCV,hong2021deep}, which are the top performers in semantic segmentation tasks, in the large-scale Cityscapes data-set and bench-marking suite for visual semantic understanding \cite{Cordts_2016_CVPR}. 

\paragraph{Dataset Acquisition}
The dataset used for training our semantic segmentation networks is generated using a combination of Unreal Engine 4 \cite{web:unrealengine} to create a realistic dock environment, and AirSim \cite{shah2018airsim} to extract raw RGB and labeled, in an automatic manner. Unreal engine is a modern {3}D engine used to create games and CGI in movies and is well known for its high-quality graphics, and AirSim a tool for simulating cars or \acp{UAV} in either Unreal Engine or Unity 3D.
To bridge the gap between the simulation and the real environment, we randomize the color, shape of the ships, weather, and time of the day (illumination and shadows) and add random Gaussian noise on the water surface shape and texture (e.g. foaming)  \cite{dehban2019impact}.
In all our experiments each network is trained from scratch using the dataset described in Table \ref{tab:semantic_dataset_description}, using \ac{SGD} and bootstrapped cross-entropy loss function \cite{bootstrappingLearning2015}, with learning rate $\eta=0.01$. To improve robustness to translations and orientations, we augment the original dataset with mirrored and flipped images. While simulation data and transfer learning can help the network learn, the training dataset contains both simulated and real-world image samples (see Fig. \ref{fig:example_labeled_images}), to incorporate smaller details that are difficult to simulate using state-of-the-art rendering engines, such as foam, smoke, surface scattering (e.g. how much the light penetrates the water). The dataset is collected at $30$ to $100$ meters altitude.

\begin{table*}[t]
\centering
\normalsize
\caption{Dataset used for training and validating the semantic segmentation networks.}
\begin{tabular}{|c|ccc|ccc|}
\hline
 \multirow{2}{*}{} & \multicolumn{3}{c|}{\textbf{Total of images in dataset}} & \multicolumn{3}{c|}{\textbf{Images containing category}} \\\cline{2-7}
      & Total & Real & Simulation & Ship & Water & Unknown \\ \hline\hline
Train & 8459  & 165  & 8294      & 8269 & 8444  & 4868   \\ \hline
Val   & 22    & 22   & 0         & 22   & 22    & 14     \\ \hline
Test  & 100   & 100  & 0         & 93   & 100   & 63     \\ \hline
\end{tabular}

\label{tab:semantic_dataset_description}
\end{table*}

\subsubsection{Image-based Distance Computation}
After finding the segmented blobs for a given image, we employ topological analysis of binarized images \cite{journals/cvgip/SuzukiA85}, to extract contours, i.e., the surrounding relations among the outermost borders 
of the binary image. The resulting silhouettes are then used for computing object-specific relative distances using two approaches described next. 

The 3D silhouette of the vessel is represented by a set of lines defined in the vessel coordinate frame $\mathcal{V}$, as follows
\begin{align}
    S_\mathcal{V}=\{ s_{v} \subset \mathcal{V} , v=1,...,N_s \}
\end{align}
is precomputed from a known CAD model and simplified using a convex hull representation, which is discretized into a user-specified number of subsections, according to safety structural requirements.
Each subsection is represented by two delimiting points, which are precomputed and stored in a list structure at initialization time. In the online step, we transform each vessel line section into the image plane coordinates, according to
\begin{align}
    i_{v}=T^\mathcal{I}_\mathcal{V} s_{v}
    \label{eq:vessel_to_pixel}
\end{align}
Then, we compute perpendicular vectors for each subsection, departing from each of these points, to define regions that enclose potential obstacles represented by non-vessel found contours. The distance between the contours and the corresponding vessel subsection is obtained by projecting each contour point $c_j$ into the line defined by the points delimiting the subsection (see Fig. \ref{fig:method_output}) and finding the minimum distance one, according to
\begin{align}
d_{v}=\text{min}_j\left( \| c_{v,j}^\perp \| \right)
\end{align}
with $c_{v,j}^\perp=c_j - i_{v,j}$, where $i_{v,j}$ is the closest point to $c_j$ on the vessel (i.e. section $v$), and $c_{v,j}^\perp$ the orthogonal vector from $i_{v,j}$ to $c_j$.

The distances in pixel coordinates are converted to distances in vessel coordinates using the \ac{GSD} function, according to
\begin{align}
D_v = \text{GSD}(d_{v}) =\text{max}\left( \frac{a\times\text{w}_s}{f\times\text{w}_i},
\frac{a\times\text{h}_s}{f\times\text{h}_i}
\right) \times d_v
\end{align}
where $a$ represents the altitude of the camera, $h_s$ the camera sensor height, $h_i$ the height, $w_s$ the camera sensor width, $w_i$ the image width and $f$ the camera focal length, $d_v$ is the min distance in pixel units and $D_v$ is the euclidean distance found in meters. This is a reasonable approximation when the viewpoint altitude is significantly larger than the height of the structures on the ground (i.e. or at sea level).


\section{Experimental Results}
\label{sec:ex_res}
The experiments intend to show the accuracy and efficiency of the proposed method as a whole. To accomplish this, we break the evaluation down into; evaluating the performance of the leading semantic segmentation networks on our dataset and the overall expected performance of the methods estimated distance measurements.

\subsection{Semantic Segmentation}
As the proposed method is developed around planar projection and \ac{GSD} to convert pixels to metric euclidean distances, every pixel matters in maintaining good accuracy, one of the significant determinants on which the method depends are the semantic segmentation models, which foremost objectives are to interpret the given input image and produce a mask separating water from solids. 
The selection of the semantic segmentation models is not only based on their accuracy, but also efficiency and speed as the pipeline might need to perform in near real-time using onboard resources. We generally have to sacrifice some accuracy to get a low memory and fast model, which is why we have chosen to test \ac{HarDNet}, \ac{BiSeNet}v2, and \ac{DDRNet}.

For the training of all models, we use a \ac{SGD} optimizer with an initial learning rate set to $0.01$, the momentum to $0.9$, and weight decay to $0.0005$. The batch size is set to $8$ per GPU, and the training is stopped after $90000$ epochs. We use Bootstrapped cross-entropy on each image instead of over a batch of images so that each image will contribute to the loss. The parameter used for bootstrapped cross-entropy is $K$ going from $100\%$ of the pixels to $20\%$, and a warm-up period from $1000$ to $9000$ with the idea of the network can learn to adapt to the easy regions first and transit to the more challenging regions.  For the rest of the parameters, the original values are being used.

The test and validation dataset only contains real-case images as we want to optimize the model for real-case data only and avoid over-fitting by training and evaluating on different domains. The test dataset is captured from a different ship, season, and time of day to determine if the models are over-fitted.

\begin{figure}[t]
    \centering
    \includegraphics[width=0.48\textwidth]{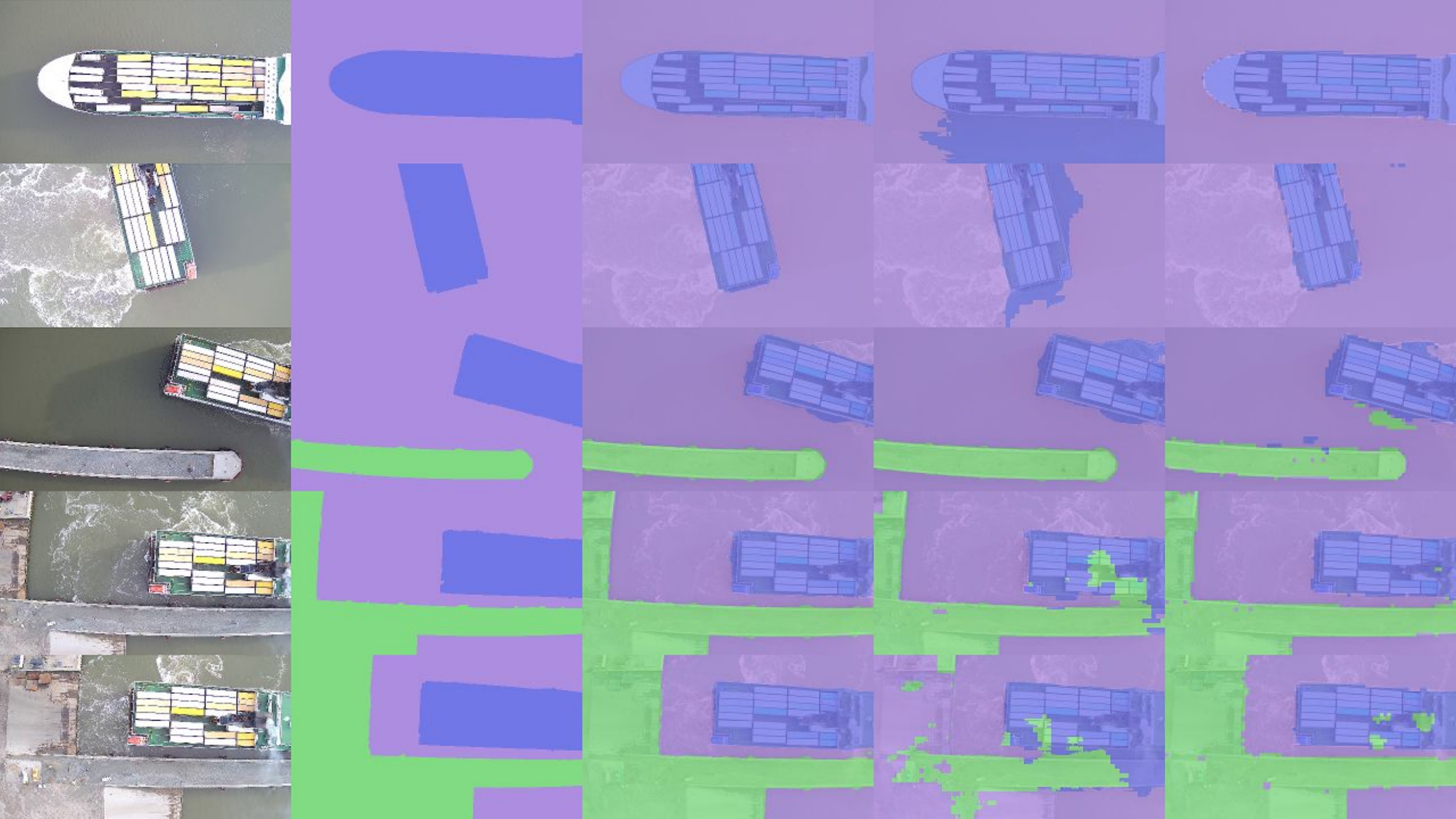}
    \caption{Segmentation image examples on the test dataset. The five columns from left to right referes to input image, ground truth label, \ac{HarDNet} prediction, \ac{BiSeNet}v2 prediction and lastly \ac{DDRNet} prediction.}
    \label{fig:showcase_of_models}
\end{figure}

Table \ref{tab:semantic_result} shows the mean accuracy of all three models on both the validation and test sets. While \ac{BiSeNet}V2 and \ac{DDRNet} are victims of overfitting, \ac{HarDNet} shows an exceptional result with high accuracy on both the validation and test set and good IoU on all three classes. This accuracy comes with the cost of being the slowest among the three models. In Fig. \ref{fig:showcase_of_models} we can see the low resolution output coming from \ac{DDRNet}. This is because the output of \ac{DDRNet} is eight times lower than the given input, which in our model is $\text{output}=80\times45$. \ac{DDRNet} is performing better on high-resolution images.
In the proposed method, only the water estimations are being processed since knowing the semantic difference of ships and harbor has little value for estimating distance on water surfaces.
\begin{table*}[t]
\centering
\normalsize
\caption{Accuracy and speed comparison on the aerial harbor dataset. The test and validation results are going in the table to reflect the amount of over-fitting in some of the networks. All neural networks are trained equally amount of epochs ($90000$) and is tested on a GTX $1080$ GPU.}
\begin{tabular}{cccccccccc}
\multirow{2}{*}{Model} &
  \multirow{2}{*}{Dataset} &
  \multirow{2}{*}{Mean accuracy} &
  \multicolumn{4}{c}{IoU} &
  \multirow{2}{*}{Speed (FPS)} &
  \multirow{2}{*}{\begin{tabular}[c]{@{}c@{}}Resolution\\ (w,h)\end{tabular}} &
  \multirow{2}{*}{Params} \\ \cline{4-7}
 &
   &
   &
  Unknown &
  Water &
  Ship &
  Mean &
   &
   &
   \\ \hline
\multirow{2}{*}{BiSeNet v2} &
  Val &
  96.4 &
  91. &
  97.7 &
  93.3 &
  94.0 &
  \multirow{2}{*}{\textbf{133.43}} &
  \multirow{2}{*}{512x288} &
  \multirow{2}{*}{3.5M} \\
 &
  Test &
  75.4 &
  47.5 &
  70.2 &
  54.5 &
  57.4 &
   &
   &
   \\ \hline
\multirow{2}{*}{HarDNet} &
  Val &
  \textbf{98.6} &
  \textbf{96.3} &
  \textbf{98.7} &
  \textbf{96.6} &
  \textbf{97.2} &
  \multirow{2}{*}{91.75} &
  \multirow{2}{*}{640x360} &
  \multirow{2}{*}{4M} \\
 &
  Test &
  \textbf{97.3} &
  \textbf{90.9} &
  \textbf{96.6} &
  \textbf{92.6} &
  \textbf{93.4} &
   &
   &
   \\ \hline
\multirow{2}{*}{DDRNet-23} &
  Val &
  97.4 &
  92.5 &
  97.6 &
  92.9 &
  94.3 &
  \multirow{2}{*}{100.12} &
  \multirow{2}{*}{640x360} &
  \multirow{2}{*}{7.6M} \\
 &
  Test &
  86.4 &
  73.4 &
  80.4 &
  62. &
  72. &
   &
   &
\end{tabular}

\label{tab:semantic_result}
\end{table*}

\subsection{Distance Estimation}
Since ground truth distance between real docking vessels and surrounding obstacles was not available during experiments, we conduct these experiments inside an AirSim simulation. We use ray-casting to compute the exact ground-truth distance between the vessel and all the other objects in the environment.
In order to evaluate the performance of the whole method, we let the \ac{UAV} move around the shipyard at different altitudes and recorded all proprioceptive and exteroceptive sensor data and ground truth. More specifically, we consider $100$ image samples taken from a fixed altitude. The number of altitudes were discretized into $a=\left[30,40,50,...,150\right]$. 

\begin{figure}[b!]
    \centering
\includegraphics[width=0.48\textwidth]{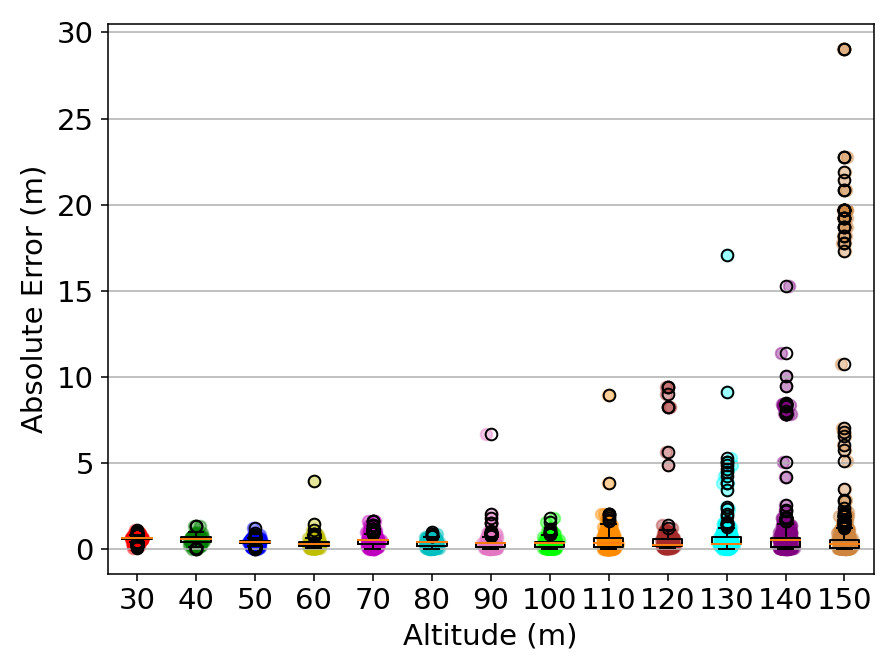}  
      \caption{The box plot shows the absolute error in relation to the \ac{UAV}'s altitude. As the altitude increases, so does the magnitude of the absolute error and the amount of outliers.}
\label{fig:bar_plot1}
\end{figure}
\begin{figure}[h!]
    \centering
\includegraphics[width=0.48\textwidth]{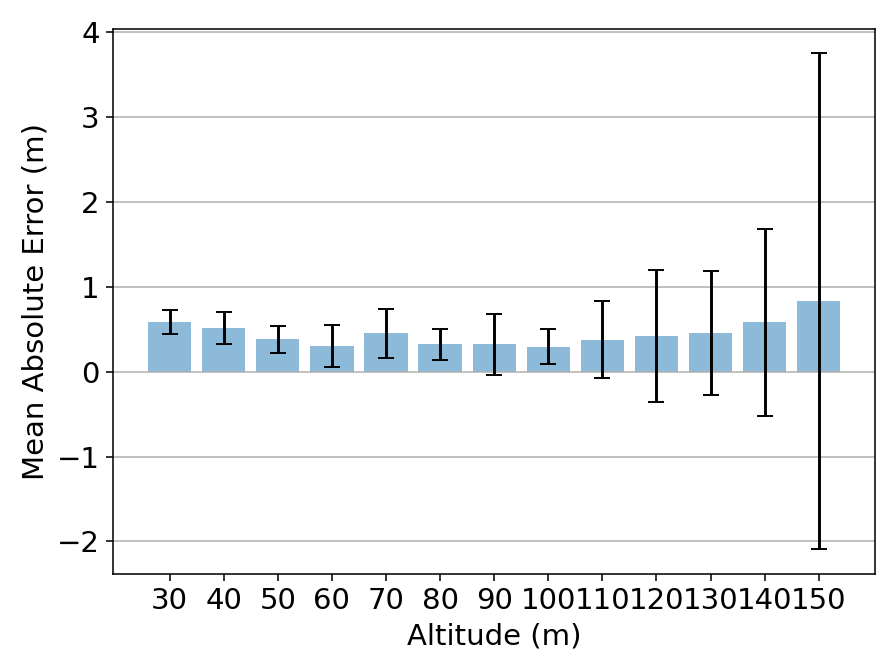}
      \caption{The blue bars represent average absolute error and black the standard deviation.}
    \label{fig:bar_plot2}
\end{figure}

Figures \ref{fig:bar_plot1} and \ref{fig:bar_plot2} show the correlation between the altitude and absolute distance error.
Figure \ref{fig:bar_plot2} is the zoomed version of Fig. \ref{fig:bar_plot1} where we can have a closer look at the mean absolute error, while Fig. \ref{fig:bar_plot2} shows the overall absolute errors of all estimated distances. The results show that the distance mean error and standard deviation increase with altitude. This is expected due to the fact that the segmentation model is only trained for up to 100 meters, which justified the amount the continuous increase in the number of outliers. Also, the \ac{GSD} error grows quadratically as altitude increases due to quantization errors.

\begin{figure}[b!]
   \centering
   \begin{subfigure}{0.22\textwidth}
     \includegraphics[width=0.98\textwidth]{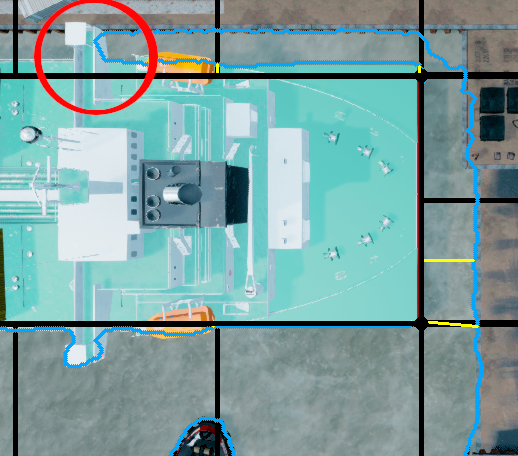}
   \end{subfigure}
   \begin{subfigure}{0.22\textwidth}
     \includegraphics[width=0.98\textwidth]{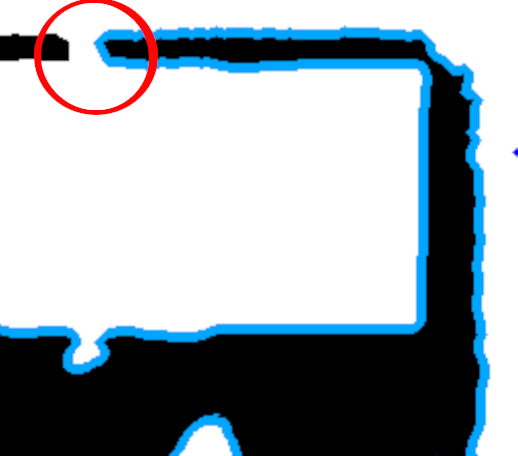}  
   \end{subfigure}        
   \caption{Example of a wrong contour due to overlap between vessel and harbor (the wide vessel bridge covers the harbor wall on left side) and the contour is wrapping around the vessel border which creates estimates of zero distance.}
   \label{fig:overlapping}
\end{figure}

\subsection{Discussion}
\label{sec:discussion}
While the average absolute error shown in Figs. \ref{fig:bar_plot1} and \ref{fig:bar_plot2} show excellent results, there are three sources of noise that contribute to errors in distance computation, depending on image acquisition altitude: The first is quantization issues. As we increase the altitude, the measuring accuracy decreases since pixel resolution is limited. The proposed method is utilizing \ac{GSD} to convert pixels into a distance metric system, and as the altitude increases, the higher value per pixel. The second is due to the planar assumption, which for lower altitudes becomes faulty. In these cases, measurement accuracy is affected by the height of the ground structures, and the extraction of contours will be faulty and noisy. This is not strictly on lower altitudes that the planar assumption is a problem. When structures are relatively high above the sea surface, they will start to overlap with the environment, and as we can see in our from Fig. \ref{fig:overlapping} at $150m$, the top of the vessel is overlapping with the surface resulting in chaotic contour results. It is a delicate balance between keeping low altitudes for better pixel accuracy and high altitude for lowering 3D disturbance. The last is due to limitations in the semantic segmentation model. 

While \ac{HarDNet} has shown the best performance in terms of accuracy, \ac{BiSeNet}, having fewer parameters, is the fastest and might be more suitable for onboard applications. As the altitude increases, the greater the distance value per pixel we get from the \ac{GSD}, and as a result, small mistakes from the model can jeopardize distance estimation accuracy.

\section{Conclusions}
In this work, we have presented a novel approach for relative inter-object distance extraction from aerial image data captured from a companion \ac{UAV}, combined with absolute \ac{GPS} and magnetometer localization data. The proposed method first employs semantic segmentation networks to discriminate between the vessel of interest and potential solid objects from raw image data, and a \ac{GSD} method that considers proprioceptive \ac{GPS} and magnetometer data to accurately determine the relative distance between the vessel and the environment obstacles. 

A dataset has been developed to train a segmentation network that consists of both photo-realistic images using AirSim and real-world images. The extensive simulation results demonstrate that the developed framework gives satisfactory accuracy in distance estimation between the object of interest (i.e. vessel) and surrounding objects. We believe the \ac{UAV}-assisted framework can improve the navigation performance of vessels when they have to maneuver in congested harbors and waterways. Although our method relies on RGB visual data for object segmentation, the proposed framework is general enough to be extended to other types of imagery data, e.g. infrared or thermal, as long as suitable training data is provided.

In future work, we intend to extend the system to fuse information over time (e.g. probabilistic Bayesian framework) to keep track of dynamic obstacles \cite{figueiredoVisapp17}, the vessel, and their distances in the environment. This way, one could extend the proposed framework for NBV planning \cite{defigueiredo2021advantages} to improve distance estimation over time and propose safe trajectories to the ship.

\section{Acknowledgement}
The authors would like to acknowledge the financial contribution from Smart Industry Program (European Regional Development Fund and Region Midtjylland, grant no.: RFM-17-0020). The authors would further like to thank Upteko Aps for bringing the use-case challenge and for collection of real life datasets.
\balance
\bibliographystyle{IEEEtran}
\bibliography{References}

\begin{thebibliography}{10}
\providecommand{\url}[1]{#1}
\csname url@samestyle\endcsname
\providecommand{\newblock}{\relax}
\providecommand{\bibinfo}[2]{#2}
\providecommand{\BIBentrySTDinterwordspacing}{\spaceskip=0pt\relax}
\providecommand{\BIBentryALTinterwordstretchfactor}{4}
\providecommand{\BIBentryALTinterwordspacing}{\spaceskip=\fontdimen2\font plus
\BIBentryALTinterwordstretchfactor\fontdimen3\font minus
  \fontdimen4\font\relax}
\providecommand{\BIBforeignlanguage}[2]{{%
\expandafter\ifx\csname l@#1\endcsname\relax
\typeout{** WARNING: IEEEtran.bst: No hyphenation pattern has been}%
\typeout{** loaded for the language `#1'. Using the pattern for}%
\typeout{** the default language instead.}%
\else
\language=\csname l@#1\endcsname
\fi
#2}}
\providecommand{\BIBdecl}{\relax}
\BIBdecl

\bibitem{Cordts_2016_CVPR}
M.~Cordts, M.~Omran, S.~Ramos, T.~Rehfeld, M.~Enzweiler, R.~Benenson,
  U.~Franke, S.~Roth, and B.~Schiele, ``The cityscapes dataset for semantic
  urban scene understanding,'' in \emph{Proceedings of the IEEE Conference on
  Computer Vision and Pattern Recognition (CVPR)}, June 2016.

\bibitem{bisenet2018}
\BIBentryALTinterwordspacing
C.~Yu, J.~Wang, C.~Peng, C.~Gao, G.~Yu, and N.~Sang, ``Bisenet: Bilateral
  segmentation network for real-time semantic segmentation,'' \emph{CoRR}, vol.
  abs/1808.00897, 2018. [Online]. Available:
  \url{http://arxiv.org/abs/1808.00897}
\BIBentrySTDinterwordspacing

\bibitem{Chao_2019_ICCV}
P.~Chao, C.-Y. Kao, Y.-S. Ruan, C.-H. Huang, and Y.-L. Lin, ``Hardnet: A low
  memory traffic network,'' in \emph{Proceedings of the IEEE/CVF International
  Conference on Computer Vision (ICCV)}, October 2019.

\bibitem{dense_networks2017}
G.~{Huang}, Z.~{Liu}, L.~{Van Der Maaten}, and K.~Q. {Weinberger}, ``Densely
  connected convolutional networks,'' in \emph{2017 IEEE Conference on Computer
  Vision and Pattern Recognition (CVPR)}, 2017, pp. 2261--2269.

\bibitem{hong2021deep}
Y.~Hong, H.~Pan, W.~Sun, and Y.~Jia, ``Deep dual-resolution networks for
  real-time and accurate semantic segmentation of road scenes,'' \emph{arXiv
  preprint arXiv:2101.06085}, 2021.

\bibitem{ilkeriros2020}
I.~{Bozcan} and E.~{Kayacan}, ``Uav-adnet: Unsupervised anomaly detection using
  deep neural networks for aerial surveillance,'' in \emph{2020 IEEE/RSJ
  International Conference on Intelligent Robots and Systems (IROS)}, Las
  Vegas, NV, USA, 2020, pp. 1158--1164.

\bibitem{bozcancontext}
------, ``Context-dependent anomaly detection for low altitude traffic
  surveillance,'' in \emph{2021 IEEE International Conference on Robotics and
  Automation (ICRA)}, Xi’an, China, To appear.

\bibitem{ecamci2020}
E.~{Camci}, D.~{Campolo}, and E.~{Kayacan}, ``Deep reinforcement learning for
  motion planning of quadrotors using raw depth images,'' in \emph{2020
  International Joint Conference on Neural Networks (IJCNN)}, 2020, pp. 1--7.

\bibitem{camci2019endtoend}
\BIBentryALTinterwordspacing
E.~Camci and E.~Kayacan, ``End-to-end motion planning of quadrotors using deep
  reinforcement learning,'' \emph{CoRR}, vol. abs/1909.13599, 2019. [Online].
  Available: \url{http://arxiv.org/abs/1909.13599}
\BIBentrySTDinterwordspacing

\bibitem{lidarCameraFusion2020}
\BIBentryALTinterwordspacing
G.~A. Kumar, J.~H. Lee, J.~Hwang, J.~Park, S.~H. Youn, and S.~Kwon, ``Lidar and
  camera fusion approach for object distance estimation in self-driving
  vehicles,'' \emph{Symmetry}, vol.~12, no.~2, 2020. [Online]. Available:
  \url{https://www.mdpi.com/2073-8994/12/2/324}
\BIBentrySTDinterwordspacing

\bibitem{measuringDistanceRadar2019}
L.~{Piotrowsky}, T.~{Jaeschke}, S.~{Kueppers}, J.~{Siska}, and N.~{Pohl},
  ``Enabling high accuracy distance measurements with fmcw radar sensors,''
  \emph{IEEE Transactions on Microwave Theory and Techniques}, vol.~67, no.~12,
  pp. 5360--5371, 2019.

\bibitem{8356033}
R.-J. Yan, E.~Kayacan, I.-M. Chen, L.~K. Tiong, and J.~Wu, ``Quicabot: Quality
  inspection and assessment robot,'' \emph{IEEE Transactions on Automation
  Science and Engineering}, vol.~16, no.~2, pp. 506--517, 2019.

\bibitem{deepLearningVehicle2020}
\BIBentryALTinterwordspacing
J.~Fayyad, M.~A. Jaradat, D.~Gruyer, and H.~Najjaran, ``Deep learning sensor
  fusion for autonomous vehicle perception and localization: A review,''
  \emph{Sensors}, vol.~20, no.~15, 2020. [Online]. Available:
  \url{https://www.mdpi.com/1424-8220/20/15/4220}
\BIBentrySTDinterwordspacing

\bibitem{ilker2021}
I.~{Bozcan}, J.~{Le Fevre}, H.~X. {Pham}, and E.~{Kayacan}, ``Gridnet:
  Image-agnostic conditional anomaly detection for indoor surveillance,''
  \emph{IEEE Robotics and Automation Letters}, vol.~6, no.~2, pp. 1638--1645,
  2021.

\bibitem{audataset2020}
I.~{Bozcan} and E.~{Kayacan}, ``Au-air: A multi-modal unmanned aerial vehicle
  dataset for low altitude traffic surveillance,'' in \emph{2020 IEEE
  International Conference on Robotics and Automation (ICRA)}, 2020, pp.
  8504--8510.

\bibitem{felipe2012analysis}
B.~Felipe-Garc{\'\i}a, D.~Hern{\'a}ndez-L{\'o}pez, and J.~L. Lerma, ``Analysis
  of the ground sample distance on large photogrammetric surveys,''
  \emph{Applied Geomatics}, vol.~4, no.~4, pp. 231--244, 2012.

\bibitem{web:unrealengine}
\BIBentryALTinterwordspacing
{Epic Games}, ``{Unreal Engine v4.24.3},'' last accessed 27 October 2020.
  [Online]. Available: \url{https://www.unrealengine.com}
\BIBentrySTDinterwordspacing

\bibitem{shah2018airsim}
S.~Shah, D.~Dey, C.~Lovett, and A.~Kapoor, ``Airsim: High-fidelity visual and
  physical simulation for autonomous vehicles,'' in \emph{Field and service
  robotics}.\hskip 1em plus 0.5em minus 0.4em\relax Springer, 2018, pp.
  621--635.

\bibitem{dehban2019impact}
A.~Dehban, J.~Borrego, R.~Figueiredo, P.~Moreno, A.~Bernardino, and
  J.~Santos-Victor, ``The impact of domain randomization on object detection: A
  case study on parametric shapes and synthetic textures,'' in \emph{2019
  IEEE/RSJ International Conference on Intelligent Robots and Systems
  (IROS)}.\hskip 1em plus 0.5em minus 0.4em\relax IEEE, 2019, pp. 2593--2600.

\bibitem{bootstrappingLearning2015}
\BIBentryALTinterwordspacing
S.~E. Reed, H.~Lee, D.~Anguelov, C.~Szegedy, D.~Erhan, and A.~Rabinovich,
  ``Training deep neural networks on noisy labels with bootstrapping,'' in
  \emph{ICLR 2015}, 2015. [Online]. Available:
  \url{http://arxiv.org/abs/1412.6596}
\BIBentrySTDinterwordspacing

\bibitem{journals/cvgip/SuzukiA85}
\BIBentryALTinterwordspacing
S.~Suzuki and K.~Abe, ``Topological structural analysis of digitized binary
  images by border following.'' \emph{Computer Vision, Graphics, and Image
  Processing}, vol.~30, no.~1, pp. 32--46, 1985. [Online]. Available:
  \url{http://dblp.uni-trier.de/db/journals/cvgip/cvgip30.html#SuzukiA85}
\BIBentrySTDinterwordspacing

\bibitem{figueiredoVisapp17}
R.~Figueiredo, J.~Avelino, A.~Dehban, A.~Bernardino, P.~Lima, and H.~Araújo,
  ``Efficient resource allocation for sparse multiple object tracking,'' in
  \emph{Proceedings of the 12th International Joint Conference on Computer
  Vision, Imaging and Computer Graphics Theory and Applications - Volume 6:
  VISAPP, (VISIGRAPP 2017)}, INSTICC.\hskip 1em plus 0.5em minus 0.4em\relax
  SciTePress, 2017, pp. 300--307.

\bibitem{defigueiredo2021advantages}
\BIBentryALTinterwordspacing
R.~P. de~Figueiredo, J.~G. Hansen, J.~L. Fevre, M.~Brand{\~{a}}o, and
  E.~Kayacan, ``On the advantages of multiple stereo vision camera designs for
  autonomous drone navigation,'' \emph{CoRR}, vol. abs/2105.12691, 2021.
  [Online]. Available: \url{https://arxiv.org/abs/2105.12691}
\BIBentrySTDinterwordspacing

\end{thebibliography}

%

\end{document}